\documentclass[11pt,a4paper]{article}
\usepackage{graphicx}
\usepackage[hyperref]{acl2017}
\usepackage{times}
\usepackage{latexsym}
\usepackage{mathtools}
\usepackage{amsmath, amsfonts}
\usepackage{nth}
\usepackage{multirow}
\usepackage{graphicx}
\usepackage{url}
\usepackage{array}
\newcolumntype{?}{!{\vrule width 1pt}}

\usepackage{mathtools}

\DeclarePairedDelimiter{\floor}{\lfloor}{\rfloor}

\aclfinalcopy 

\title{Neobility at SemEval-2017 Task 1: An Attention-based Sentence Similarity Model}

\author{WenLi Zhuang \thanks{\ \ The author is currently serving in his Alternative Military Service of Education.} \\
	Shan-Si Elementary School  \\
    ChangHua County, Taiwan \\
  {\tt bibo9901@gmail.com} \\\And
  Ernie Chang \\
  Department of Linguistics \\
  University of Washington \\
  Seattle, WA 98195, USA \\
  {\tt cyc025@uw.edu} \\}

\date{}

\begin{document}
\maketitle
\begin{abstract}
This paper describes a neural-network model which performed competitively (top 6) at the SemEval 2017 cross-lingual Semantic Textual Similarity (STS) task. Our system employs an attention-based recurrent neural network model that optimizes the sentence similarity. In this paper, we describe our participation in the multilingual STS task which measures similarity across English, Spanish, and Arabic. 
\end{abstract}

\section{Introduction}

Semantic textual similarity (STS) measures the degree
of equivalence between the meanings of two
text sequences \cite{agirre-EtAl:2016:SemEval1}. The similarity
of the text pair can be represented as
discrete or continuous values ranging from irrelevance (1) to exact
semantic equivalence (5). It is widely applicable to many NLP tasks including summarization \cite{wong2008extractive,nenkova2011automatic}, generation and question answering \cite{vo2015fbk}, paraphrase detection \cite{fernando2008semantic}, and machine translation \cite{corley2005measuring}. 

In this paper, we describe a system that is able to learn context-sensitive features within the sentences. Further, we encode the sequential information with Recurrent Neural Network (RNN) and perform attention mechanism \cite{bahdanau2014neural} on RNN outputs for both sentences. Attention mechanism was performed to increase sensitivity of the system to words of similarity significance. We also optimize directly on the Pearson correlation score as part of our neural-based approach. Moreover, we include a pair feature adapter module that could be used to include more features to further improve performance. However, for this competition we include merely the TakeLab features \cite{vsaric-EtAl:2012:STARSEM-SEMEVAL}. \footnote{Our system and data is available at \texttt{https://github.com/iamalbert/semval2017task1}.}

\section{Related Works}

Most proposed approaches in the past adopted a hybrid of varying text unit sizes ranging from character-based, token-based, to knowledge-based similarity measure \cite{gomaa2013survey}. The linguistic depths of these measures often vary between lexical, syntactic, and semantic levels. 

Most solutions include an ensemble of modules that employs features coming from different unit sizes and depths. More recent approaches generally include the word embedding-based similarity \cite{liebeck-EtAl:2016:SemEval,brychcin-svoboda:2016:SemEval} as a component of the final ensemble. Top performing team in 2016  \cite{Rychalska-EtAl:2016:SemEval} uses an ensemble of multiple modules including recursive autoencoders with WordNet and monolingual aligner \cite{Sultan-EtAl:2016:SemEval}. UMD-TTIC-UW \cite{he-EtAl:2016:SemEval2} proposes the MPCNN model that requires no feature engineering and managed to perform competitively at 6\textsuperscript{th} place. MPCNN is able to extract the hidden information using the Convolutional Neural Network (CNN) and added an attention layer to extract the vital similarity information. 

\section{Methods}
\subsection{Model}
\begin{figure*}
\includegraphics[width=\linewidth]{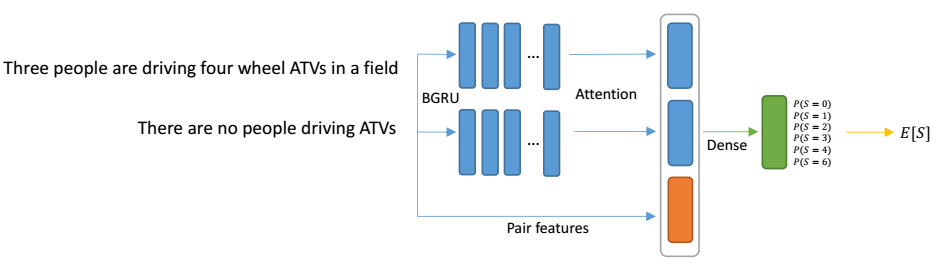}
\caption{Illustration of model architecture}
\label{fig:model}
\end{figure*}

Given two sentences $I_1 = \lbrace w^1_1, w^1_2, ..., w^1_{n^1} \rbrace$ and $I_2 = \lbrace w^2_1, w^2_2, ..., w^2_{n^2} \rbrace$, where $w_{ij}$ denote the $j$th token of the $i$th sentence, embedded using a function $\phi$ that maps each token to a $D$-dimension trainable vector. Two sentences are encoded with an attentitve RNN to obtain sentence embeddings $u^1$, $u^2$, respectively.
 \paragraph{Sentence Encoder} For each sentence, the RNN firstly converts $w^i_{j}$ into $x^i_{j} \in R^{2H}$, using an bidirectional Gated Recurrent Unit (GRU) \cite{cho-EtAl:2014:EMNLP2014} \footnote{ We also explored Longer Short-Term Memory (LSTM), but find it more overfitting than GRU.} by sequentially fed $w^i_j$ into the unit, forward and backward. The superscripts of $w,x,a,u,n$ are omitted for clear notation. 
\begin{equation}
\begin{aligned}
x_{i} &= [ x^F_{i}; x^B_{i} ] \\
x^F_{i} &= \text{GRU}(x^F_{i-1}, w_{i}) \\
x^B_{i} &= \text{GRU}(x^F_{i+1}, w_{i})
\end{aligned}
\end{equation}
Then, we attend each word $x_j$ for different salience $a_j$ and blend the moemories $x_{1;n}$ into sentence embedding $u$:
\begin{equation}
\begin{aligned}
a_j &\propto \exp( r^T\tanh(W x_i) ) \\
u & = \sum_{j=1}^n a_j x_j 
\end{aligned}
\end{equation}

\paragraph{Surface Features}
Inspired by the \textit{simple} system described in \cite{takelab:2016:SemEval},
We also extract surface features from the sentence pair as following:

\begin{description}

	\item[$\bullet$Ngram Overlap Similarity:] 
These are features drawn from external knowledge like WordNet \cite{miller1995wordnet} and Wikipedia. We use both PathLen similarity \cite{leacock1998combining} and Lin similarity \cite{lin1998information} to compute similarity between pairs of words $w^1_i$ and $w^2_j$ in $I_1$ and $I_2$, respectively. We employed the suggested preprocessing step \cite{takelab:2016:SemEval}, and added both WordNet and corpus-based information to ngram overlap scores, which was obtained with the harmonic mean of the degree of overlap between the sentences. 

	\item[$\bullet$Semantic Sentence Similarity:] We also computed token-based alignment overlap and vector space sentence similarity \cite{vsaric-EtAl:2012:STARSEM-SEMEVAL}. Semantic alignment similarity was computed greedily between all pairs of tokens using both the knowledge-based and corpus-based similarity. Scores are further enhanced with the aligned pair information. We obtained the weighted form of latent semantic analysis vectors \cite{turney2010frequency} for each word \textit{w}, before computing the cosine similarity. As such, sentence similarity scores are enhanced with corpus-based information for tokens. The features are concatenated into a vector, denoted as $m$.

\end{description}

\paragraph{Scoring}
Let $S$ be a descret random variable over $\{0,1,2,3,4,5\}$ describing the similarity of the given sentence pair $\{I_1, I_2\}$. The representation of the given pair is the concatenation of $u^1$, $u^2$, and $m$, which is fed into an MLP with one hidden layer to calculate the estimated distribution of $S$.
\begin{equation}
	p=\left[ 
    	\begin{array}{cc} P(S=0) \\ P(S=1) \\ \vdots \\ P(S=5) \end{array}
    \right]
    = 
    \text{softmax}( V \tanh (U \left[
    	\begin{array}{cc} u^1\\u^2\\m \end{array}
    \right    ]))
\end{equation}
Therefore, the score $y$ is the expected value of $S$: 
\begin{equation}
	y=E[S]=\sum_{i=0}^{5} iP(S=i) = v^Tp
\end{equation}, where $v=[0,1,2,3,4,5]^T$. The entire system is shown in  Figure \ref{fig:model}.

\subsection{Word Embedding}
We explored initializing word embeddings randomly or with pre-trained word2vec \cite{mikolov2013distributed} of dimension 50, 100, 300, respectively. We found that the system works the best with 300-dimension word2vec embeddings.

\subsection{Optimization}
Let $p^n,y^n$ be the predicted probability density and expected score and $\hat{y}^n $ be the annotated gold score of the $n$-th sample. Most of the previous learning-based models are trained to minimize the following objectives on a batch of $N$ samples:
\begin{itemize}
\item Negative Log-likelihood (NLL) of $p$ and $\hat{p}$ \cite{aker-EtAl:2016:SemEval}. The task is viewed as a classification problem for 6 classes.
\begin{equation*}
L_\text{NLL} = \sum_{n=1}^N - \log p^n_{t^n}
\end{equation*}, where $t^n$ is $\hat{y}^n$ rounded to the nearest integer.
\item Mean square error (MSE) between $y^n$ and $\hat{y}^n$ \cite{brychcin-svoboda:2016:SemEval}. 
\begin{equation*}
L_\text{MSE} = \frac{1}{N} \sum_{n=1}^N ( y^n - \hat{y}^n )^2
\end{equation*}
\item Kullback-Leibler divergence (KLD) of $p^n$ and gold distribution $\hat{p}^n$ estimated by $\hat{y}^n$:
\begin{equation*}
 L_\text{KLD} = \sum_{n=1}^N \left( \sum_{i=1}^6 
		\hat{p}^n_i \log{\frac{\hat{p}^n_i}{p^n_i}} \right)
\end{equation*} where 
\begin{equation*}
	\hat{p}^n_i = \left\lbrace \begin{array}{ll}
    	\hat{y}^n - \floor*{\hat{y}^n} ,& \mbox{if}~i = \floor*{\hat{y}^n} + 1 \\
        \floor*{\hat{y}^n} + 1 - \hat{y}^n , & \mbox{if}~i = \floor{\hat{y}^n} \\
        0, & \mbox{otherwise}
    	\end{array}
     	\right.
\end{equation*}\cite{S16-1088,tai-socher-manning:2015:ACL-IJCNLP}. For each $n$, there exists some $k$ such that $\hat{p}^n_k=1$ and $\forall h \ne k, \hat{p}^n_h = 0 $, KLD is identical to NLL.
\end{itemize}

However, the evaluation metric of this task is Pearson Correlation Coefficient (PCC), which is invariant to changes in location and scale of $y^n$ but none of the above objectives can reflect it. Here we use an example to illustrate that MSE and KLD can even report an inverse tendency. In Table \ref{table:labelprob}, group A has lower MSE and KLD loss than group B, but its PCC is also lower.

To solve this problem, we train the model to maximize PCC directly. Hence, the loss function is given by:
\begin{equation}
L_\text{PCC}= -\frac{
	\sum_{n=1}^N (y^n - \bar{y})(\hat{y}^n - \bar{\hat{y}})
}{
	\sqrt[]{\sum_{n=1}^N (y^n - \bar{y})^2} ~
    \sqrt[]{\sum_{n=1}^N (\hat{y}^n - \bar{\hat{y}})^2}
}
\end{equation}
where $\bar{y} = \frac{1}{N} \sum_{n=1}^N y^n $ and $\bar{\hat{y}} = \frac{1}{N} \sum_{n=1}^N \hat{y}^n $.
Since $N$ is fixed for every batch, $L_\text{PCC}$ is differentiable with respect to $y^n$, which means we can apply back propagation to train the network. To the best of our knowledge, we are the first team to adopt this training objective.
\begin{table}[!htb]
\centering
\resizebox{\linewidth}{!}{\begin{tabular}{llll?lll}
Group      & A           &              &             & B           &              &             \\ \hline
Gold Score & 3           & 4            & 5           & 3           & 4            & 5           \\ \hline
$P(S=0)$   & 0.05        & 0.05         & 0.05        & 0.15        & 0.05         & 0.1         \\
$P(S=1)$   & 0.05        & 0.05         & 0.05        & 0.3         & 0.2          & 0.1        \\
$P(S=2)$   & 0.15        & 0.1          & 0.05        & 0.25        & 0.3          & 0.2       \\
$P(S=3)$   & 0.5         & 0.35         & 0.0         & 0.1         & 0.25         & 0.3       \\
$P(S=4)$   & 0.15        & 0.4          & 0.1         & 0.1         & 0.1          & 0.2       \\
$P(S=5)$   & 0.1         & 0.05         & 0.7         & 0.1         & 0.1          & 0.1        \\
$E[S]$     & 2.95        & 3.15         & 4.2         & 2.0         & 2.45         & 2.7       \\ \hline
           & MSE         & KLD          & PCC                  & MSE      & KLD           & PCC           \\
           & \textbf{0.455}    & \textbf{1.966}   & 0.931      & 2.90 & 6.91 & \textbf{0.987}
\end{tabular}}
                                 \caption{ Example of lower MSE and KLD not indicating higher PCC.  }
\label{table:labelprob}
\end{table}

\section{Evaluation}
\subsection{Data}
\begin{table}[!htb]
\begin{center}
\begin{tabular}{|c|c|c|}
\hline \bf Dataset & \bf Pairs \\ \hline
 Training &  22,401 \\
 Validation &  5,601  \\
\hline
\end{tabular}
\end{center}
\caption{\label{font-table} Training and validation Data sets (STS 2012-2016 and SICK). }
\label{table:dataset}
\end{table}
We gathered dataset from SICK \cite{MARELLI14.363} and past STS across years 2012, 2013, 2014, 2015, and 2016 \cite{agirre-EtAl:2012:STARSEM-SEMEVAL,agirre-EtAl:2013:*SEM1,agirre-EtAl:2014:SemEval,agirre-EtAl:2015:SemEval2,agirre-EtAl:2016:SemEval1} for both cross-lingual and monolingual subtasks. We shuffling and splitting them according to the ratio 80:20 into training set and validation set, respectively. Table \ref{table:dataset} indicates the size of training set and validation set. All non-English sentence appearing in training, validation, and test set are translated into English with Google Cloud Translation API.
\subsection{Experiments}

In the experiment, the size of output of GRU is set to be $H=200$. We use ADAM algorithm to optimize the parameters with mini-batches of 125. The learning rate is set to $10^{-4}$ at the beginning and reduced by half for every 5 epochs. We trained the network for 15 epochs.

\paragraph{Word embeddings}
In Table \ref{table:wordemb}, we demonstrate that the system performs better with pretrained word vectors (WI) than randomly initialized (RI).
\begin{table}[h]
\centering
\begin{tabular}{|c|c|c|}
\hline
                    & $D$ & PCC on validation set \\ \hline
\multirow{2}{*}{RI} & 50                   & 0.7904         \\ \cline{2-3} 
                    & 300                  & 0.8091         \\ \hline
\multirow{2}{*}{WI} & 50                   & 0.7974         \\ \cline{2-3} 
                    & 300                  & 0.\textbf{8174}         \\ \hline
\end{tabular}
\caption{System performance with different dimensions of word embeddings, using either randomly initialized or pre-trained word embedding.}
\label{table:wordemb}
\end{table}

\begin{figure}
\begin{center}
\includegraphics[width=\linewidth]{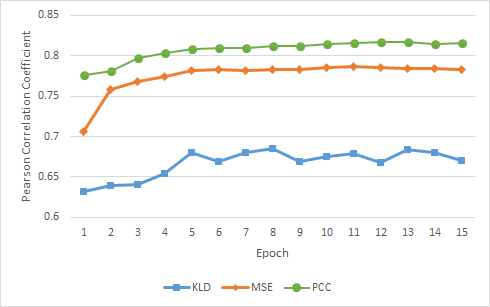}
\end{center}
\caption{Performance of different loss functions}
\label{fig:losses}
\end{figure}

\paragraph{Loss function} We display performances with systems optimized with KLD, MSE, and PCC. It shows that when using $L_\text{PCC}$ as the training objective, our system not only performs the best but also converges the fastest. As shown in Table \ref{table:loss} and Figure \ref{fig:losses}.
\begin{table}[h]
\centering
\begin{tabular}{|c|c|c|}
\hline Loss function & PCC \\ \hline
	$L_\text{KLD}$ &  0.6839 \\
    $L_\text{MSE}$ &  0.7863 \\
    $L_\text{PCC}$ &  \textbf{0.8174} \\
\hline
\end{tabular}
\caption{Influence of different loss objectives on the system performance measured using PCC on our validation set.}
\label{table:loss}
\end{table}

\subsection{Final System Results}
We tune the model on validation set, and select the set of hyper-parameters that yields the best performance to obtain the scores of test data. We report the official provisional results in Table \ref{table:final}.
There is an obvious performance drop in track4b, which happens to all teams. We hypothesized that the sentences in track4b (en\_es) are collected from a special domain, due to the fact that the number of out-of-vocabulary words in track 4b is many times more than that in other tracks.

\begin{table}[h]
\centering
\begin{tabular}{|c|c||ccc|}
\hline Track & PCC &mean & median & max \\ \hline
Primary & 0.6171 & 0.66 & 0 & 28 \\ \hline
	1   & 0.6821 & 0.53 & 0 & 3 \\
    2   & 0.6459 & 0.50 & 0 & 3 \\
    3   & 0.7928 & 0.35 & 0 & 4 \\
    4a  & 0.7169 & 0.35 & 0 & 4 \\
    4b  & 0.0200 & 2.54 & 2 & 28 \\
    5   & 0.7927 & 0.36 & 0 & 4 \\
    6   & 0.6696 & 0.33 & 0 & 5 \\ \hline
\end{tabular}
\caption{Final system results and statistics of the number of OOV words within a pair}
\label{table:final}
\end{table}

\section{Conclusion and Future Work}

In conclusion, we propose a simple neural-based system with a novel means of optimization. We found that optimizing directly on PCC achieved the best scores, allowing the model to perform competitively on STS-2017. Moreover, we demonstrated that using randomly initialized word embedding does not harm the performance, but allowing it to achieve slightly higher scores against the pre-trained word embedding.





\bibliography{acl2017}

\begin{thebibliography}{}
\expandafter\ifx\csname natexlab\endcsname\relax\def\natexlab#1{#1}\fi

\bibitem[{Agirre et~al.(2015)Agirre, Banea, Cardie, Cer, Diab, Gonzalez-Agirre,
  Guo, Lopez-Gazpio, Maritxalar, Mihalcea, Rigau, Uria, and
  Wiebe}]{agirre-EtAl:2015:SemEval2}
Eneko Agirre, Carmen Banea, Claire Cardie, Daniel Cer, Mona Diab, Aitor
  Gonzalez-Agirre, Weiwei Guo, Inigo Lopez-Gazpio, Montse Maritxalar, Rada
  Mihalcea, German Rigau, Larraitz Uria, and Janyce Wiebe. 2015.
\newblock \href{http://www.aclweb.org/anthology/S15-2045}{Semeval-2015 task 2:
  Semantic textual similarity, english, spanish and pilot on interpretability}.
\newblock In {\em Proceedings of the 9th International Workshop on Semantic
  Evaluation (SemEval 2015)\/}. Association for Computational Linguistics,
  Denver, Colorado, pages 252--263.
\newblock
  \href{http://www.aclweb.org/anthology/S15-2045}{http://www.aclweb.org/anthology/S15-2045}.

\bibitem[{Agirre et~al.(2014)Agirre, Banea, Cardie, Cer, Diab, Gonzalez-Agirre,
  Guo, Mihalcea, Rigau, and Wiebe}]{agirre-EtAl:2014:SemEval}
Eneko Agirre, Carmen Banea, Claire Cardie, Daniel Cer, Mona Diab, Aitor
  Gonzalez-Agirre, Weiwei Guo, Rada Mihalcea, German Rigau, and Janyce Wiebe.
  2014.
\newblock \href{http://www.aclweb.org/anthology/S14-2010}{Semeval-2014 task 10:
  Multilingual semantic textual similarity}.
\newblock In {\em Proceedings of the 8th International Workshop on Semantic
  Evaluation (SemEval 2014)\/}. Association for Computational Linguistics and
  Dublin City University, Dublin, Ireland, pages 81--91.
\newblock
  \href{http://www.aclweb.org/anthology/S14-2010}{http://www.aclweb.org/anthology/S14-2010}.

\bibitem[{Agirre et~al.(2016)Agirre, Banea, Cer, Diab, Gonzalez-Agirre,
  Mihalcea, Rigau, and Wiebe}]{agirre-EtAl:2016:SemEval1}
Eneko Agirre, Carmen Banea, Daniel Cer, Mona Diab, Aitor Gonzalez-Agirre, Rada
  Mihalcea, German Rigau, and Janyce Wiebe. 2016.
\newblock \href{http://www.aclweb.org/anthology/S16-1081}{Semeval-2016 task 1:
  Semantic textual similarity, monolingual and cross-lingual evaluation}.
\newblock In {\em Proceedings of the 10th International Workshop on Semantic
  Evaluation\/}. Association for Computational Linguistics, San Diego,
  California, pages 497--511.
\newblock
  \href{http://www.aclweb.org/anthology/S16-1081}{http://www.aclweb.org/anthology/S16-1081}.

\bibitem[{Agirre et~al.(2012)Agirre, Cer, Diab, and
  Gonzalez-Agirre}]{agirre-EtAl:2012:STARSEM-SEMEVAL}
Eneko Agirre, Daniel Cer, Mona Diab, and Aitor Gonzalez-Agirre. 2012.
\newblock \href{http://www.aclweb.org/anthology/S12-1051}{Semeval-2012 task 6:
  A pilot on semantic textual similarity}.
\newblock In {\em {*SEM 2012}: The First Joint Conference on Lexical and
  Computational Semantics -- Volume 1: Proceedings of the main conference and
  the shared task, and Volume 2: Proceedings of the Sixth International
  Workshop on Semantic Evaluation {(SemEval 2012)}\/}. Association for
  Computational Linguistics, Montr\'{e}al, Canada, pages 385--393.
\newblock
  \href{http://www.aclweb.org/anthology/S12-1051}{http://www.aclweb.org/anthology/S12-1051}.

\bibitem[{Agirre et~al.(2013)Agirre, Cer, Diab, Gonzalez-Agirre, and
  Guo}]{agirre-EtAl:2013:*SEM1}
Eneko Agirre, Daniel Cer, Mona Diab, Aitor Gonzalez-Agirre, and Weiwei Guo.
  2013.
\newblock \href{http://www.aclweb.org/anthology/S13-1004}{*sem 2013 shared
  task: Semantic textual similarity}.
\newblock In {\em Second Joint Conference on Lexical and Computational
  Semantics (*SEM), Volume 1: Proceedings of the Main Conference and the Shared
  Task: Semantic Textual Similarity\/}. Association for Computational
  Linguistics, Atlanta, Georgia, USA, pages 32--43.
\newblock
  \href{http://www.aclweb.org/anthology/S13-1004}{http://www.aclweb.org/anthology/S13-1004}.

\bibitem[{Aker et~al.(2016)Aker, Blain, Duque, Fomicheva, Seva, Shah, and
  Beck}]{aker-EtAl:2016:SemEval}
Ahmet Aker, Frederic Blain, Andres Duque, Marina Fomicheva, Jurica Seva, Kashif
  Shah, and Daniel Beck. 2016.
\newblock \href{http://www.aclweb.org/anthology/S16-1092}{Usfd at semeval-2016
  task 1: Putting different state-of-the-arts into a box}.
\newblock In {\em Proceedings of the 10th International Workshop on Semantic
  Evaluation (SemEval-2016)\/}. Association for Computational Linguistics, San
  Diego, California, pages 609--613.
\newblock
  \href{http://www.aclweb.org/anthology/S16-1092}{http://www.aclweb.org/anthology/S16-1092}.

\bibitem[{Bahdanau et~al.(2015)Bahdanau, Cho, and Bengio}]{bahdanau2014neural}
Dzmitry Bahdanau, Kyunghyun Cho, and Yoshua Bengio. 2015.
\newblock Neural machine translation by jointly learning to align and
  translate.
\newblock In {\em Proc. {ICLR}\/}.

\bibitem[{Brychc\'{i}n and Svoboda(2016)}]{brychcin-svoboda:2016:SemEval}
Tom\'{a}\v{s} Brychc\'{i}n and Luk\'{a}\v{s} Svoboda. 2016.
\newblock \href{http://www.aclweb.org/anthology/S16-1089}{Uwb at semeval-2016
  task 1: Semantic textual similarity using lexical, syntactic, and semantic
  information}.
\newblock In {\em Proceedings of the 10th International Workshop on Semantic
  Evaluation (SemEval-2016)\/}. Association for Computational Linguistics, San
  Diego, California, pages 588--594.
\newblock
  \href{http://www.aclweb.org/anthology/S16-1089}{http://www.aclweb.org/anthology/S16-1089}.

\bibitem[{Cho et~al.(2014)Cho, van Merrienboer, Gulcehre, Bahdanau, Bougares,
  Schwenk, and Bengio}]{cho-EtAl:2014:EMNLP2014}
Kyunghyun Cho, Bart van Merrienboer, Caglar Gulcehre, Dzmitry Bahdanau, Fethi
  Bougares, Holger Schwenk, and Yoshua Bengio. 2014.
\newblock \href{http://www.aclweb.org/anthology/D14-1179}{Learning phrase
  representations using rnn encoder--decoder for statistical machine
  translation}.
\newblock In {\em Proceedings of the 2014 Conference on Empirical Methods in
  Natural Language Processing (EMNLP)\/}. Association for Computational
  Linguistics, Doha, Qatar, pages 1724--1734.
\newblock
  \href{http://www.aclweb.org/anthology/D14-1179}{http://www.aclweb.org/anthology/D14-1179}.

\bibitem[{Corley and Mihalcea(2005)}]{corley2005measuring}
Courtney Corley and Rada Mihalcea. 2005.
\newblock Measuring the semantic similarity of texts.
\newblock In {\em Proceedings of the ACL workshop on empirical modeling of
  semantic equivalence and entailment\/}. Association for Computational
  Linguistics, pages 13--18.

\bibitem[{Fernando and Stevenson(2008)}]{fernando2008semantic}
Samuel Fernando and Mark Stevenson. 2008.
\newblock A semantic similarity approach to paraphrase detection.
\newblock In {\em Proceedings of the 11th Annual Research Colloquium of the UK
  Special Interest Group for Computational Linguistics\/}. Citeseer, pages
  45--52.

\bibitem[{Gomaa and Fahmy(2013)}]{gomaa2013survey}
Wael~H Gomaa and Aly~A Fahmy. 2013.
\newblock A survey of text similarity approaches.
\newblock {\em International Journal of Computer Applications\/} 68(13).

\bibitem[{He et~al.(2016)He, Wieting, Gimpel, Rao, and
  Lin}]{he-EtAl:2016:SemEval2}
Hua He, John Wieting, Kevin Gimpel, Jinfeng Rao, and Jimmy Lin. 2016.
\newblock \href{http://www.aclweb.org/anthology/S16-1170}{Umd-ttic-uw at
  semeval-2016 task 1: Attention-based multi-perspective convolutional neural
  networks for textual similarity measurement}.
\newblock In {\em Proceedings of the 10th International Workshop on Semantic
  Evaluation (SemEval-2016)\/}. Association for Computational Linguistics, San
  Diego, California, pages 1103--1108.
\newblock
  \href{http://www.aclweb.org/anthology/S16-1170}{http://www.aclweb.org/anthology/S16-1170}.

\bibitem[{Leacock and Chodorow(1998)}]{leacock1998combining}
Claudia Leacock and Martin Chodorow. 1998.
\newblock Combining local context and wordnet similarity for word sense
  identification.
\newblock {\em WordNet: An electronic lexical database\/} 49(2):265--283.

\bibitem[{Li and Huang(2016)}]{S16-1088}
Peng Li and Heng Huang. 2016.
\newblock \href{https://doi.org/10.18653/v1/S16-1088}{Uta dlnlp at semeval-2016
  task 1: Semantic textual similarity: A unified framework for semantic
  processing and evaluation}.
\newblock In {\em Proceedings of the 10th International Workshop on Semantic
  Evaluation (SemEval-2016)\/}. Association for Computational Linguistics,
  pages 584--587.
\newblock
  \href{https://doi.org/10.18653/v1/S16-1088}{https://doi.org/10.18653/v1/S16-1088}.

\bibitem[{Liebeck et~al.(2016)Liebeck, Pollack, Modaresi, and
  Conrad}]{liebeck-EtAl:2016:SemEval}
Matthias Liebeck, Philipp Pollack, Pashutan Modaresi, and Stefan Conrad. 2016.
\newblock \href{http://www.aclweb.org/anthology/S16-1090}{Hhu at semeval-2016
  task 1: Multiple approaches to measuring semantic textual similarity}.
\newblock In {\em Proceedings of the 10th International Workshop on Semantic
  Evaluation (SemEval-2016)\/}. Association for Computational Linguistics, San
  Diego, California, pages 595--601.
\newblock
  \href{http://www.aclweb.org/anthology/S16-1090}{http://www.aclweb.org/anthology/S16-1090}.

\bibitem[{Lin et~al.(1998)}]{lin1998information}
Dekang Lin et~al. 1998.
\newblock An information-theoretic definition of similarity.
\newblock In {\em ICML\/}. Citeseer, volume~98, pages 296--304.

\bibitem[{Marelli et~al.(2014)Marelli, Menini, Baroni, Bentivogli, Bernardi,
  and Zamparelli}]{MARELLI14.363}
Marco Marelli, Stefano Menini, Marco Baroni, Luisa Bentivogli, Raffaella
  Bernardi, and Roberto Zamparelli. 2014.
\newblock A sick cure for the evaluation of compositional distributional
  semantic models.
\newblock In Nicoletta Calzolari~(Conference Chair), Khalid Choukri, Thierry
  Declerck, Hrafn Loftsson, Bente Maegaard, Joseph Mariani, Asuncion Moreno,
  Jan Odijk, and Stelios Piperidis, editors, {\em Proceedings of the Ninth
  International Conference on Language Resources and Evaluation (LREC'14)\/}.
  European Language Resources Association (ELRA), Reykjavik, Iceland.

\bibitem[{Mikolov et~al.(2013)Mikolov, Sutskever, Chen, Corrado, and
  Dean}]{mikolov2013distributed}
Tomas Mikolov, Ilya Sutskever, Kai Chen, Greg~S Corrado, and Jeff Dean. 2013.
\newblock Distributed representations of words and phrases and their
  compositionality.
\newblock In {\em Advances in neural information processing systems\/}. pages
  3111--3119.

\bibitem[{Miller(1995)}]{miller1995wordnet}
George~A Miller. 1995.
\newblock Wordnet: a lexical database for english.
\newblock {\em Communications of the ACM\/} 38(11):39--41.

\bibitem[{Nenkova et~al.(2011)Nenkova, McKeown et~al.}]{nenkova2011automatic}
Ani Nenkova, Kathleen McKeown, et~al. 2011.
\newblock Automatic summarization.
\newblock {\em Foundations and Trends{\textregistered} in Information
  Retrieval\/} 5(2--3):103--233.

\bibitem[{Rychalska et~al.(2016)Rychalska, Pakulska, Chodorowska, Walczak, and
  Andruszkiewicz}]{Rychalska-EtAl:2016:SemEval}
Barbara Rychalska, Katarzyna Pakulska, Krystyna Chodorowska, Wojciech Walczak,
  and Piotr Andruszkiewicz. 2016.
\newblock \href{https://doi.org/10.18653/v1/S16-1091}{Samsung poland nlp team
  at semeval-2016 task 1: Necessity for diversity; combining recursive
  autoencoders, wordnet and ensemble methods to measure semantic similarity.}
\newblock In {\em Proceedings of the 10th International Workshop on Semantic
  Evaluation (SemEval-2016)\/}. Association for Computational Linguistics,
  pages 602--608.
\newblock
  \href{https://doi.org/10.18653/v1/S16-1091}{https://doi.org/10.18653/v1/S16-1091}.

\bibitem[{{\v{S}}ari{\'{c}} et~al.(2012){\v{S}}ari{\'{c}}, Glava{\v{s}}, Karan,
  {\v{S}}najder, and Dalbelo~Ba{\v{s}}i{\'{c}}}]{takelab:2016:SemEval}
Frane {\v{S}}ari{\'{c}}, Goran Glava{\v{s}}, Mladen Karan, Jan {\v{S}}najder,
  and Bojana Dalbelo~Ba{\v{s}}i{\'{c}}. 2012.
\newblock \href{http://aclweb.org/anthology/S12-1060}{Takelab: Systems for
  measuring semantic text similarity}.
\newblock In {\em *SEM 2012: The First Joint Conference on Lexical and
  Computational Semantics -- Volume 1: Proceedings of the main conference and
  the shared task, and Volume 2: Proceedings of the Sixth International
  Workshop on Semantic Evaluation (SemEval 2012)\/}. Association for
  Computational Linguistics, pages 441--448.
\newblock
  \href{http://aclweb.org/anthology/S12-1060}{http://aclweb.org/anthology/S12-1060}.

\bibitem[{Sultan et~al.(2016)Sultan, Bethard, and
  Sumner}]{Sultan-EtAl:2016:SemEval}
Arafat~Md Sultan, Steven Bethard, and Tamara Sumner. 2016.
\newblock \href{https://doi.org/10.18653/v1/S16-1099}{Dls\$@\$cu at
  semeval-2016 task 1: Supervised models of sentence similarity}.
\newblock In {\em Proceedings of the 10th International Workshop on Semantic
  Evaluation (SemEval-2016)\/}. Association for Computational Linguistics,
  pages 650--655.
\newblock
  \href{https://doi.org/10.18653/v1/S16-1099}{https://doi.org/10.18653/v1/S16-1099}.

\bibitem[{Tai et~al.(2015)Tai, Socher, and
  Manning}]{tai-socher-manning:2015:ACL-IJCNLP}
Kai~Sheng Tai, Richard Socher, and Christopher~D. Manning. 2015.
\newblock \href{http://www.aclweb.org/anthology/P15-1150}{Improved semantic
  representations from tree-structured long short-term memory networks}.
\newblock In {\em Proceedings of the 53rd Annual Meeting of the Association for
  Computational Linguistics and the 7th International Joint Conference on
  Natural Language Processing (Volume 1: Long Papers)\/}. Association for
  Computational Linguistics, Beijing, China, pages 1556--1566.
\newblock
  \href{http://www.aclweb.org/anthology/P15-1150}{http://www.aclweb.org/anthology/P15-1150}.

\bibitem[{Turney and Pantel(2010)}]{turney2010frequency}
Peter~D Turney and Patrick Pantel. 2010.
\newblock From frequency to meaning: Vector space models of semantics.
\newblock {\em Journal of artificial intelligence research\/} 37:141--188.

\bibitem[{Vo et~al.(2015)Vo, Magnolini, and Popescu}]{vo2015fbk}
Ngoc Phuoc~An Vo, Simone Magnolini, and Octavian Popescu. 2015.
\newblock Fbk-hlt: An application of semantic textual similarity for answer
  selection in community question answering.
\newblock In {\em Proceedings of the 9th International Workshop on Semantic
  Evaluation, SemEval\/}. volume~15, pages 231--235.

\bibitem[{\v{S}ari\'{c} et~al.(2012)\v{S}ari\'{c}, Glava\v{s}, Karan,
  \v{S}najder, and Dalbelo~Ba\v{s}i\'{c}}]{vsaric-EtAl:2012:STARSEM-SEMEVAL}
Frane \v{S}ari\'{c}, Goran Glava\v{s}, Mladen Karan, Jan \v{S}najder, and
  Bojana Dalbelo~Ba\v{s}i\'{c}. 2012.
\newblock \href{http://www.aclweb.org/anthology/S12-1060}{Takelab: Systems for
  measuring semantic text similarity}.
\newblock In {\em {*SEM 2012}: The First Joint Conference on Lexical and
  Computational Semantics -- Volume 1: Proceedings of the main conference and
  the shared task, and Volume 2: Proceedings of the Sixth International
  Workshop on Semantic Evaluation {(SemEval 2012)}\/}. Association for
  Computational Linguistics, Montr\'{e}al, Canada, pages 441--448.
\newblock
  \href{http://www.aclweb.org/anthology/S12-1060}{http://www.aclweb.org/anthology/S12-1060}.

\bibitem[{Wong et~al.(2008)Wong, Wu, and Li}]{wong2008extractive}
Kam-Fai Wong, Mingli Wu, and Wenjie Li. 2008.
\newblock Extractive summarization using supervised and semi-supervised
  learning.
\newblock In {\em Proceedings of the 22nd International Conference on
  Computational Linguistics-Volume 1\/}. Association for Computational
  Linguistics, pages 985--992.

\end{thebibliography}
\bibliographystyle{acl_natbib}

\appendix

\end{document}